\documentclass[runningheads]{llncs}
\usepackage[T1]{fontenc}
\usepackage{graphicx}
\usepackage{caption}
\usepackage{multirow}
\usepackage{booktabs}
\usepackage{graphicx}
\usepackage{subcaption}
\usepackage{amsmath}
\usepackage{amssymb}
\usepackage{hyperref}
\usepackage{pgfplots}
\usepackage{pgfplotstable}
\usepackage{svg}

\usepackage{siunitx}
\usepackage{makecell}

\usepackage{bm}
\usepackage[ruled,vlined]{algorithm2e}

\usepgfplotslibrary{statistics}

\usepgfplotslibrary{groupplots,fillbetween}
\pgfplotsset{compat=1.18}

\newcommand{\orcid}[1]{\href{https://orcid.org/#1}{\includegraphics[width=0.03\textwidth]{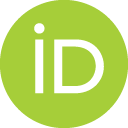}}}

\begin{document}
\title{End-to-End Keyword Spotting on FPGA Using Graph Neural Networks with a Neuromorphic Auditory Sensor} 
\titlerunning{End-to-End Keyword Spotting on FPGA Using GNNs and NAS}

\author{Wiktor Matykiewicz\inst{1}\orcid{0009-0005-2648-4415} \and
Piotr Wzorek\inst{1,2}\orcid{0000-0003-3885-600X} \and
Kamil Jeziorek\inst{1,2}*\orcid{0000-0001-5446-3682} \and
Tomás Muñoz\inst{3} \and
Antonio	Rios-Navarro\inst{3}\orcid{0000-0003-4163-8484} \and
Angel Jiménez-Fernández\inst{3}\orcid{0000-0003-3061-5922} \and \\
Tomasz Kryjak\inst{1, 2}\orcid{0000-0001-6798-4444}
} 


\institute{AGH University of Krakow, Poland \and
Embedded Vision Systems Group, Computer Vision Laboratory \and
Robotics and Technology of Computers Lab., ETSII, EPS, SCORE, I3US, Universidad de Sevilla\\
* Corresponding author: kjeziorek@agh.edu.pl
}

\authorrunning{W. Matykiewicz et al}

%
\maketitle              
\begin{abstract}
With the rapid growth of mobile robotics and embedded intelligence, there is an increasing demand for efficient on-device data processing on edge platforms. 
A promising research direction is the use of neuromorphic sensors inspired by human sensory systems, which generate sparse, event-based data encoding changes in the environment. 
In this work, we present the first end-to-end FPGA implementation of a keyword spotting system that integrates a Neuromorphic Auditory Sensor (NAS) and a graph neural network (GNN) on a single FPGA device, enabling real-time processing of raw audio data. 
The proposed architecture eliminates conventional signal preprocessing and operates directly on event-based audio streams. 
Leveraging a compute-near-memory network architecture, the system achieves efficient inference with low latency and low power consumption. 
Experimental results demonstrate an accuracy of 87.43\% after quantization on the Google Speech Commands v2 dataset processed through the neuromorphic sensor, with end-to-end latency below 35 µs and average power consumption of 1.12 W. 
The processed datasets, software models, and hardware modules are available at \url{https://github.com/vision-agh/NAS-GNN-KWS}.

\end{abstract}
\section{Introduction}

Edge perception has become a key requirement in modern embedded systems, including mobile robotics and IoT applications. 
In such scenarios, audio processing plays an important role and must ensure high prediction accuracy while maintaining low energy consumption and minimal latency. 
In this context, neuromorphic sensors are gaining increasing attention, as they generate events representing changes in the environment. 
The resulting data are inherently sparse in both space and time, and preserving this sparsity during processing enables a substantial reduction in computational complexity.  

An example of such a device for audio processing is the Neuromorphic Auditory Sensor (NAS)~\cite{jimenez2017binaural}, whose FPGA-based implementation enables direct conversion of a raw digital audio stream into event-based representations. Among the methods capable of efficiently processing sparse event data, spiking neural networks and graph neural networks (GNN) have been widely investigated. In~\cite{Jeziorek2026HAGNN}, an FPGA-based architecture for keyword spotting (KWS) utilizing GNN and operating on simulated event-based data was proposed, demonstrating low latency and reduced energy consumption.  

In this work, our main objective is to demonstrate the feasibility of integrating the NAS and a GNN within a single FPGA device, forming a unified end-to-end system suitable for mobile robotics applications. The main contributions of this paper are as follows:
\begin{itemize}
    \item We present and publicly release a set of event-based audio datasets recorded using multiple NAS configurations. The datasets correspond to the popular Google Speech Commands v2 benchmark \cite{warden2018speech} for keyword spotting.
    \item We provide detailed statistical analysis of the recorded data and propose an event filtering method that reduces the number of processed events by approximately 47\%, while simultaneously improving prediction accuracy. The method is designed for efficient hardware implementation.
    \item We introduce a set of architectural modifications to the GNN, derived from extensive ablation studies, enabling integration of the sensor and network on a single device while ensuring sufficient throughput. The proposed changes improve latency while reducing energy consumption.
    \item We present the first end-to-end KWS system operating on real event-based data from a neuromorphic auditory sensor, integrating both the sensor and a graph neural network on the same FPGA device. The system is evaluated in simulation and on the target hardware platform, achieving high accuracy, low latency, and low energy consumption.
\end{itemize}

The remainder of this paper is organized as follows. Section \ref{sec:realated} reviews related work. Section \ref{sec:proposed_method} describes the proposed method. Section \ref{sec:evaluation} presents the experimental evaluation. Finally, Section \ref{sec:summary} concludes the paper.

\section{Related work}
\label{sec:realated}
\subsection{Neuromorphic Auditory Sensors}

Biological auditory systems operate asynchronously, generating sparse spikes only when spectral energy changes occur—unlike conventional pipelines that rely on fixed-rate sampling and computationally intensive DSP. This event-driven principle significantly reduces redundant data transmission and enables efficient processing of dynamic acoustic environments. Such bio-inspired systems offer the high dynamic range and low power consumption required for edge applications.

Early hardware realizations employed analog VLSI circuits to mimic cochlear functionality~\cite{lyonandmead,chan2007aer,wang2015design,DAS_arch}, but such systems suffer from transistor mismatch and noise, and key platforms are no longer commercially available. Digital alternatives on FPGAs offer greater scalability, though conventional filter-bank implementations retain significant arithmetic overhead~\cite{gambin2010digital,thakur2014fpga}. In this work, we adopt the spike signal processing (SSP) paradigm~\cite{5596845} using the Neuromorphic Auditory Sensor (NAS)~\cite{jimenez2017binaural}.  This architecture processes information directly in the spike domain, avoiding conventional numerical signal representations. By replacing complex arithmetic operations with simple logic gates and counters, SSP enables efficient event-based computation, allowing for fully parallel auditory processing with minimal power consumption.

\subsection{Processing event-based audio data}

Event-based audio from NAS is sparse, asynchronous, and temporally irregular, posing challenges for standard DSP pipelines. Learning-based methods therefore either operate directly on the event stream or convert events into temporally accumulated representations suitable for tensor-based processing.

Spiking neural networks (SNNs) remain a central modelling paradigm for event-based audio due to their native compatibility with spike-domain inputs. Existing work includes convolutional~\cite{rossbroich2022,hammouamri2023,Sadovsky2023} and recurrent~\cite{cramer2020,perez2021,Yin2021,dampfhoffer2022,bittar2022} architectures. More recently, state-space models (SSMs) have been investigated as an alternative framework for event-driven sequences~\cite{schone2024,huber2024scaling}. SSMs provide a structured mechanism for representing evolving latent states over time and can capture long-range temporal dependencies without relying on explicit recurrence.

Another emerging direction is the use of graph neural networks (GNNs), where event streams are interpreted as collections of interacting elements rather than uniformly sampled time series~\cite{lars,nakano2025hardware}.
By representing events as nodes and encoding relationships through edges, these methods can express locality and interaction patterns that are difficult to encode with fixed-grid representations. Work \cite{Jeziorek2026HAGNN} extends this approach by integrating GNNs with recurrence to increase temporal dependency over longer contexts, which is the baseline for our study.

Alongside these algorithmic developments, a number of recent studies have focused on efficient hardware realization of event-driven processing pipelines, particularly on FPGA platforms~\cite{carpegna2024spiker,quantisenc2024,nakano2025hardware,Jeziorek2026HAGNN}. Event-driven computation aligns well with FPGA architectures because sparse activity can reduce unnecessary switching and memory operations, improving energy efficiency and latency.

A recurring limitation of prior work is the reliance on synthetically generated event streams. Benchmarks such as SHD and SSC~\cite{cramer_heidelberg_2020} are frequently used for SNN evaluation but they may not fully capture the variability, noise characteristics, and timing irregularities present in real NAS outputs. In contrast, our system processes event-based audio from a physical neuromorphic sensor integrated with a GNN on a single FPGA, alongside an efficient hardware realisation of the full processing pipeline.

\section{The proposed method}
\label{sec:proposed_method}

The main goal of this work is to build an end-to-end keyword spotting system on a single FPGA device that processes raw audio data from a neuromorphic auditory sensor.
Our system consists of three main components: the neuromorphic sensor that generates event-based audio data from raw audio input (Section~\ref{subsec:sensor}), the filtration stage that reduces the number of events (Section~\ref{subsec:filtration}), and the graph generation and GNN architecture that processes the event-based data (Section~\ref{subsec:gnn_model}).
In this section, we describe the implementation of each component and the overall architecture of the system.


\subsection{Sensor}
\label{subsec:sensor}

The input stage of our system utilizes a fully digital FPGA-based NAS architecture following the spike signal processing principles described in \cite{5596845} and \cite{jimenez2017binaural}. This module is responsible for converting continuous audio signals into a sparse, asynchronous stream of Address-Event Representation (AER) packets.

The signal acquisition pipeline begins with an analog audio input provided via a standard jack interface. An external audio codec digitizes the analog signal and transmits it to the FPGA via an I2S digital audio interface. Inside the FPGA, the pulse coded modulation (PCM) samples are immediately converted into a high-frequency spike train using pulse frequency modulation (PFM). This spike train serves as the input to a bank of digital spike-based filters that decompose the signal into distinct frequency bands (channels).

To investigate the trade-offs between biological plausibility, latency, and event throughput, we implemented and evaluated the NAS in two distinct topological configurations:

\begin{itemize}
    \item \textbf{Cascade Architecture:} Inspired by the biological cochlea and \cite{jimenez2017binaural}, this configuration chains spike low-pass filters (SLPFs), each feeding into the next to progressively remove higher frequencies. However, cumulative latency scales with the number of channels, and non-ideal filter charachteristics cause compounding signal attenuation, reducing the event rate.
    
    \item \textbf{Parallel Architecture:} Here, the input spike train feeds simultaneously into a bank of independent spike band-pass filters (SBPFs). This eliminates cumulative delay and avoids compounding signal degradation, yielding similar latency across frequency channels and higher event density (Table~\ref{tab:statistics}).
    \end{itemize}

The output of the filter banks is transmitted via an asynchronous AER interface. Natively, the sensor encodes each event as a tuple $(c, p)$, representing the frequency channel index and signal polarity, respectively. To incorporate temporal information, we designed and implemented a custom timestamping module on the FPGA. This module assigns a timestamp $t$ with a resolution of $1 \, \mu s$ to each event immediately upon generation, yielding the final event representation $(t, c, p)$ used for subsequent processing.

We synthesized and tested the system in six different variations to analyze the impact of spectral resolution on KWS performance. Specifically, we implemented both cascade and parallel architectures for 32, 64, and 128 frequency channels. Increasing the number of channels improves frequency resolution but results in higher FPGA resource utilization (Slices and LUTs) and increased power consumption. The resulting event streams undergo a filtration stage before serving as the input to our graph neural network pipeline.

\pgfplotstableread[col sep=comma]{figures/statistics/events_per_channel_stats_32_serial.csv}\TserialA
\pgfplotstableread[col sep=comma]{figures/statistics/events_per_channel_stats_32_parallel.csv}\TparA

\pgfplotstableread[col sep=comma]{figures/statistics/events_per_channel_stats_64_serial.csv}\TserialB
\pgfplotstableread[col sep=comma]{figures/statistics/events_per_channel_stats_64_parallel.csv}\TparB

\pgfplotstableread[col sep=comma]{figures/statistics/events_per_channel_stats_128_serial.csv}\TserialC
\pgfplotstableread[col sep=comma]{figures/statistics/events_per_channel_stats_128_parallel.csv}\TparC

\begin{figure}[!t]
\centering
\captionsetup{type=table}
\caption{Statistics of the generated neuromorphic version of the Google Speech Commands v2 dataset~\cite{warden2018}. Reported values are the mean $\pm$ standard deviation and the maximum number of events, expressed in kilo-events (kEv.).}
\label{tab:statistics}

\setlength{\tabcolsep}{5pt}
\resizebox{\columnwidth}{!}{%
\begin{tabular}{@{}lcccccc@{}}
\toprule
\textbf{Version} &
\textbf{Max. kEv./sample} &
\textbf{Avg. kEv./sample} &
\textbf{Max. kEv./s} &
\textbf{Avg. kEv./s} \\
\midrule
32-cascade    & 204.84 & $17.22 \pm 9.58$  & 151.21 & $15.03 \pm 8.21$ \\
32-parallel  & 227.22 & $39.19 \pm 12.65$  & 174.82 & $28.08 \pm 9.12$ \\
64-cascade    & 227.99 & $31.67 \pm 13.96$  & 194.64 & $24.63 \pm 11.15$ \\
64-parallel  & 245.35 & $75.40 \pm 21.41$  & 277.07 & $54.01 \pm 16.14$ \\
128-cascade   & 254.67 & $61.27 \pm 20.94$  & 240.57 & $45.49 \pm 16.01$ \\
128-parallel & 265.96 & $126.69 \pm 24.11$ & 518.69 & $92.18 \pm 21.03$ \\
\bottomrule
\end{tabular}%
}

\vspace{1.0ex}

\captionsetup{type=figure}

\begin{tikzpicture}
\definecolor{SerialCol}{RGB}{0,114,178}
\definecolor{ParCol}{RGB}{213,94,0}

\begin{groupplot}[
    group style={
        group size=1 by 3,
        vertical sep=4.2ex,
        x descriptions at=edge bottom,
        y descriptions at=edge left,
        xticklabels at=all,
    },
    width=0.75\columnwidth,
    height=0.125\columnwidth,
    scale only axis,
    grid=both,
    tick align=outside,
    tick pos=left,
    axis y line*=left,
    ymin=0,
    enlarge x limits=false,
    enlarge y limits=false,
    xlabel={Channel index},
    ylabel={Avg. \#events},
    label style={font=\small},
    ylabel style={font=\small},
    ticklabel style={font=\small},
    yticklabel style={font=\small, text=SerialCol},
    x tick label style={/pgf/number format/fixed, yshift=1pt},
    legend style={draw=none, fill=none, font=\small, at={(0.02,0.98)}, anchor=north west},
    legend columns=2,
]

\nextgroupplot[xmin=-1, xmax=33, ymax=1750, xtick={0,8,16,24,32}]
\addlegendimage{thick, mark=none, color=SerialCol}
\addlegendentry{cascade}
\addlegendimage{thick, mark=none, color=ParCol}
\addlegendentry{parallel}

\node[anchor=north east, font=\small\bfseries,
      fill=white, fill opacity=0.8, text opacity=1, rounded corners=1pt, inner sep=1pt]
    at (axis description cs:0.98,0.98) {(a) 32 ch};

\addplot[thick, mark=none, color=SerialCol, name path=meanS32]
    table[x=channel, y=mean] {\TserialA};

\addplot[draw=none, name path=upperS32, forget plot]
    table[x=channel, y expr=\thisrow{mean}+\thisrow{std}] {\TserialA};
\addplot[draw=none, name path=lowerS32, forget plot]
    table[x=channel, y expr=\thisrow{mean}-\thisrow{std}] {\TserialA};

\addplot[fill=SerialCol, fill opacity=0.18, draw=none, forget plot]
    fill between[of=upperS32 and lowerS32];

\nextgroupplot[xmin=-2, xmax=66, xtick={0,16,32,48,64}]
\node[anchor=north east, font=\small\bfseries,
      fill=white, fill opacity=0.8, text opacity=1, rounded corners=1pt, inner sep=1pt]
    at (axis description cs:0.98,0.98) {(b) 64 ch};

\addplot[thick, mark=none, color=SerialCol, name path=meanS64]
    table[x=channel, y=mean] {\TserialB};

\addplot[draw=none, name path=upperS64, forget plot]
    table[x=channel, y expr=\thisrow{mean}+\thisrow{std}] {\TserialB};
\addplot[draw=none, name path=lowerS64, forget plot]
    table[x=channel, y expr=\thisrow{mean}-\thisrow{std}] {\TserialB};

\addplot[fill=SerialCol, fill opacity=0.18, draw=none, forget plot]
    fill between[of=upperS64 and lowerS64];

\nextgroupplot[xmin=-4, xmax=132, xtick={0,32,64,96,128}]
\node[anchor=north east, font=\small\bfseries,
      fill=white, fill opacity=0.8, text opacity=1, rounded corners=1pt, inner sep=1pt]
    at (axis description cs:0.98,0.98) {(c) 128 ch};

\addplot[thick, mark=none, color=SerialCol, name path=meanS128]
    table[x=channel, y=mean] {\TserialC};

\addplot[draw=none, name path=upperS128, forget plot]
    table[x=channel, y expr=\thisrow{mean}+\thisrow{std}] {\TserialC};
\addplot[draw=none, name path=lowerS128, forget plot]
    table[x=channel, y expr=\thisrow{mean}-\thisrow{std}] {\TserialC};

\addplot[fill=SerialCol, fill opacity=0.18, draw=none, forget plot]
    fill between[of=upperS128 and lowerS128];

\end{groupplot}

\begin{groupplot}[
    group style={
        group size=1 by 3,
        vertical sep=4.2ex,
        x descriptions at=edge bottom,
        y descriptions at=edge right,
        xticklabels at=all,
    },
    width=0.75\columnwidth,
    height=0.125\columnwidth,
    scale only axis,
    grid=none,
    tick align=outside,
    tick pos=right,
    axis y line*=right,
    ymin=0,
    enlarge x limits=false,
    enlarge y limits=false,
    axis x line=none,
    xmajorticks=false,
    xmin=0,
    xlabel={},
    label style={font=\small},
    ylabel style={font=\small, text=ParCol},
    ticklabel style={font=\small},
    yticklabel style={font=\small, text=ParCol},
    legend style={draw=none, fill=none},
]

\nextgroupplot[xmin=-1, xmax=33]
\addplot[thick, mark=none, color=ParCol, name path=meanP32]
    table[x=channel, y=mean] {\TparA};

\addplot[draw=none, name path=upperP32, forget plot]
    table[x=channel, y expr=\thisrow{mean}+\thisrow{std}] {\TparA};
\addplot[draw=none, name path=lowerP32, forget plot]
    table[x=channel, y expr=\thisrow{mean}-\thisrow{std}] {\TparA};

\addplot[fill=ParCol, fill opacity=0.12, draw=none, forget plot]
    fill between[of=upperP32 and lowerP32];

\nextgroupplot[xmin=-2, xmax=66]
\addplot[thick, mark=none, color=ParCol, name path=meanP64]
    table[x=channel, y=mean] {\TparB};

\addplot[draw=none, name path=upperP64, forget plot]
    table[x=channel, y expr=\thisrow{mean}+\thisrow{std}] {\TparB};
\addplot[draw=none, name path=lowerP64, forget plot]
    table[x=channel, y expr=\thisrow{mean}-\thisrow{std}] {\TparB};

\addplot[fill=ParCol, fill opacity=0.12, draw=none, forget plot]
    fill between[of=upperP64 and lowerP64];

\nextgroupplot[xmin=-4, xmax=132]
\addplot[thick, mark=none, color=ParCol, name path=meanP128]
    table[x=channel, y=mean] {\TparC};
\addplot[draw=none, name path=upperP128, forget plot]
    table[x=channel, y expr=\thisrow{mean}+\thisrow{std}] {\TparC};
\addplot[draw=none, name path=lowerP128, forget plot]
    table[x=channel, y expr=\thisrow{mean}-\thisrow{std}] {\TparC};
\addplot[fill=ParCol, fill opacity=0.12, draw=none, forget plot]
    fill between[of=upperP128 and lowerP128];

\end{groupplot}

\end{tikzpicture}

\caption{Average events per channel with standard deviation for different configurations of the neuromorphic auditory sensor. For all configurations the number of events is higher for low channel (higher frequency) indices.}
\label{fig:events_per_channel}
\end{figure}

\subsection{Filtration}
\label{subsec:filtration}

As shown in Section~\ref{subsec:sensor}, the sensor produces a large number of events (Table~\ref{tab:statistics}), and their distribution is highly uneven across channels (Fig.~\ref{fig:events_per_channel}). 
This variability increases the computational cost of subsequent processing and can bias the graph construction toward channels with the highest activity. 
Therefore, before graph generation, we apply a dedicated filtration stage that reduces the event rate.

The proposed filtration is designed to mimic the behavior of a leaky integrate-and-fire (LIF) neuron model \cite{abbott1999lapicque}. 
Each channel maintains an internal state that integrates incoming events as a discrete potential and decays over time; once the potential exceeds a channel-specific threshold, the event is accepted and the potential is reset. 

In our implementation, the filtration is controlled by three parameters: \texttt{div\_\allowbreak factor}, which defines the time quantization used for the decay; the weight $w$ added to potential; and the per-channel thresholds ${\theta_c}$. 
The overall procedure is summarized in Algorithm~\ref{alg:filtration}, which performs a per-channel update for every event and outputs a filtered event set $\mathcal{P}$.

A key element of the filtration is the selection of per-channel thresholds ${\theta_c}$. 
To generate these thresholds, we consider three simple strategies: constant, linear, and exponential, whose effect on the event reduction and downstream performance is evaluated in the Section~\ref{subsec:filtration_ablation}.

\begin{algorithm}[!t]
\caption{Per-channel filtration with decayed potential}
\label{alg:filtration}
\scriptsize
\DontPrintSemicolon
\KwIn{Event stream $\{(t_i, c_i, p_i)\}_{i=1}^{N}$ with time $t_i\in\mathbb{N}$, channel $c_i \in \{0,\dots,C-1\}$ and polarity $p_i \in \{-1, 1\}$}
\KwIn{Parameters: $\texttt{div\_factor}\in\mathbb{N}$, weight $w\ge 0$, thresholds $\{\theta_c\}_{c=0}^{C-1}$}
\KwOut{Accepted events $\mathcal{P}$ (each as $(t,c, p)$)}

\BlankLine
\textbf{State:} per-channel last timestamp $t_{\text{last}}[c]$ and potential $v[c]$\;

\BlankLine
\textbf{Initialization:}\;
\For{$c\leftarrow 0$ \KwTo $C-1$}{
    $t_{\text{last}}[c] \leftarrow 0$\;
    $v[c] \leftarrow 0$\;
}
$\mathcal{P} \leftarrow \emptyset$\;
$q \leftarrow 2^{\texttt{div\_factor}}$\tcp*{time quantization for decay}

\BlankLine
\For{$i\leftarrow 1$ \KwTo $N$}{
    $t \leftarrow t_i$;\quad $c \leftarrow c_i$;\quad $p \leftarrow p_i$\;
    $\Delta t \leftarrow t - t_{\text{last}}[c]$\;
    $v[c] \leftarrow \max\!\Bigl(0,\, v[c] - \Bigl\lfloor \dfrac{\Delta t}{q} \Bigr\rfloor\Bigr) + w$\tcp*{decay + integrate}
    $t_{\text{last}}[c] \leftarrow t$\;

    \If{$v[c] < \theta_c$}{
        \textbf{continue}\
    }
    $\mathcal{P} \leftarrow \mathcal{P} \cup \{(t,c, p)\}$\ \tcp*{accept event and reset potential}
    $v[c] \leftarrow 0$\;
}
\Return{$\mathcal{P}$}\;
\end{algorithm}


\subsection{Graph Generation and GNN architecture}
\label{subsec:gnn_model}

In our work, we adopted the FPGA-based hardware implementation proposed in~\cite{Jeziorek2026HAGNN} (for detailed description refer to full paper), designed for the keyword spotting task using event-driven audio data. The architecture follows a data-flow paradigm, where consecutive modules process the recorded event streams.

In the considered system, each input event is first forwarded to the graph generation module, where it is represented as a vertex. For every newly registered event, directed edges are created toward previously recorded events (time-directed graph) that lie within a half-sphere defined by the search radius along the channel dimension ($R_{\mathrm{c}}$) and the temporal dimension ($R_{t}$ - we support both the lower time radius $R_t^{low}$ and the upper time radius $R_t^{high}$). For this work, the values of these parameters were determined through ablation studies (see Section~\ref{subsec:ablation_studies}). This design enables sparse, event-driven processing immediately after acquisition and dynamic graph updates. 

The resulting vertex, together with its edge list, is subsequently processed by the feature extraction stage composed of four consecutive \texttt{PointNetConv} modules~\cite{qi2017pointnet}. The operation of \texttt{PointNetConv} can be expressed as:
\begin{equation}
\mathbf{x}'_i = \left( \max_{j \in \mathcal{N}(i) \cup \{ i \}} 
\phi_{\Theta} \left( \mathbf{x}_j, \mathbf{p}_j - \mathbf{p}_i \right) \right),
\end{equation}
where $\mathbf{x}_i$ denotes the input feature vector of node $i$, $\mathbf{p}_i$ its position, $\mathcal{N}(i)$ the neighborhood of node $i$, and $\phi_{\Theta}$ is a learnable function.  

For the first \texttt{PointNetConv} layer, two input features are used, corresponding to the mean neighbor position (channel and timestamp). The ablation studies were conducted to select the number of output features in each \texttt{PointNetConv} module in a way that guaranties efficient BRAM utilization, enabling full use of the memory word width.

The subsequent module is a MaxPool layer that aggregates events within 10\,ms temporal windows. After this interval, the aggregated feature vector is forwarded to the network head, which consists of four linear layers and GRU-based memory unit. Every 10\,ms the GNN generates prediction consisting of:
\begin{itemize}
    \item \textbf{class}: For the keyword spotting task, we selected a set of target keywords and an additional \textit{unknown} class that includes all remaining words, silence, and noise. The number of outputs depends on the number of selected keywords.
    \item \textbf{conf}: An additional single-valued output representing the confidence that a keyword has been detected within a given time window. This end-of-word temporal localization significantly improves the practical performance.
\end{itemize}

In~\cite{Jeziorek2026HAGNN}, the system was evaluated on the Spiking Heidelberg Digits and Spiking Speech Commands datasets~\cite{cramer_heidelberg_2020}, which are synthetic spiking datasets obtained through simulation (with 700 audio channels each). In this work, we introduced a series of modifications to enable deployment on data recorded directly by the NAS on the same target device. To this end, we defined system-level requirements and proposed necessary optimizations.

Due to the targeted application in mobile robotics and edge processing, we identified energy consumption and system latency as the key design constraints.  
The proposed system must therefore exhibit low resource utilization. To further minimize energy consumption, we implemented the entire processing pipeline exclusively in reconfigurable logic, avoiding the programmable processing system of the heterogeneous platform. The raw audio signal can be acquired directly on the FPGA and processed up to prediction without CPU involvement.  

At the same time, the graph neural network must ensure sufficient system throughput. To determine precise performance requirements, we analyzed the event rate after filtering (see. Section \ref{subsec:gnn_model}). The system must sustain real-time processing even during periods of increased activity, which are often critical for accurate keyword spotting.

\subsubsection{Performed modifications.}
\label{subsec:modification}

The hardware architecture was designed to support all considered configurations.
The NAS provides an extra attribute for each event: polarity, which we incorporated as an additional input feature to the first convolutional layer.
To enable deployment on the target platform while minimizing energy consumption—without violating latency and throughput constraints—we optimized the feature extraction module of the GNN.  

\begin{figure}[!t]
    \centering
    \resizebox{\textwidth}{!}{
    \includegraphics{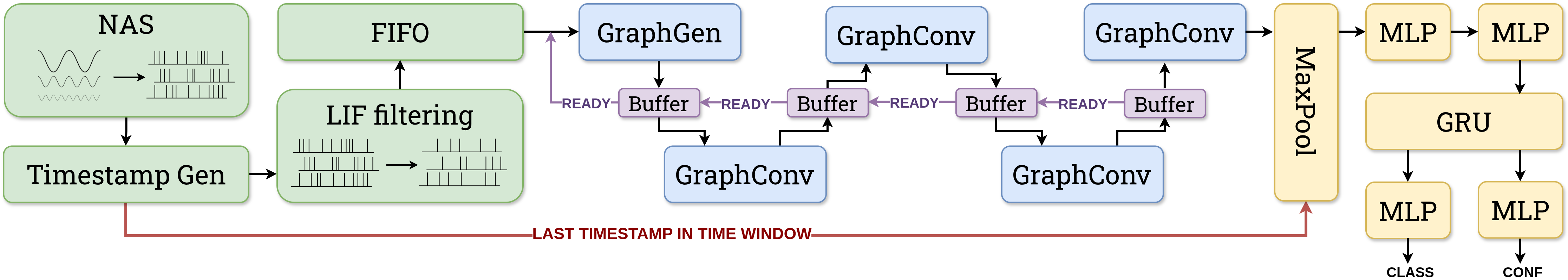}
    }
    \caption{The proposed architecture is illustrated with the sensor and filtering modules highlighted in green, the feature extraction stage in blue, and the MaxPool and network head modules in yellow. The scheduling mechanism is marked in purple, while the timestamp propagation mechanism is indicated in red.}
    \label{fig:design}
\end{figure}


LUT utilization in considered architecture is dominated by vector multipliers, which are the core computational elements of the \texttt{PointNetConv} layers. While~\cite{Jeziorek2026HAGNN} employed an architecture supporting four parallel feature-vector multiplications per graph convolution (74 8-bit elements each), we reduced the number of parallel multipliers to decrease resource usage. Since graph convolution forms the system bottleneck, increasing sequential processing impacts both latency and throughput. The final number of multipliers was selected based on the measured average and maximum post-filter event rates for each NAS configuration, defining explicit requirements (see Table~\ref{tab:all_best_configs}).  

Moreover, the computational load of a single \texttt{PointNetConv} operation depends on the number of edges, as linear transformations must be applied to the processed vertex and each of its neighbors. In the~\cite{Jeziorek2026HAGNN} design, input events were accepted at a fixed interval (in clock cycles) determined by the worst-case neighborhood size, ensuring no internal congestion. This conservative approach resulted in constant (yet limited) throughput and unnecessary latency.  

We therefore introduced a back-pressure scheduling mechanism with inter-module buffering (see Fig.~\ref{fig:design}). Each processing stage is equipped with a local buffer and signals readiness upon completing the current convolution. As the latency of each operation depends on the actual number of neighbors, events are admitted dynamically once downstream resources become available. Consequently, latency becomes event-dependent, and the input rate adapts to instantaneous workload conditions. Despite reducing the number of parallel multipliers, the effective throughput satisfies system requirements.

It is important to note that introducing the scheduling mechanism disrupts a fixed throughput. In the original design, the MaxPool module relied on a constant convolution latency, triggering feature extraction every 10\,ms based on a real-time counter. After our modifications, it became necessary to implement a \textit{timestamp propagation mechanism}, in which the NAS output timestamps are monitored and the last timestamp within each time window is propagated to the MaxPool module. The MaxPool then forwards features to the network head only when the final event of the corresponding 10\,ms window has been processed.

\section{Evaluation}
\label{sec:evaluation}

This section evaluates the proposed keyword spotting system in terms of both classification performance and end-of-word temporal localization.
We first describe the training and evaluation setup and define the metrics used throughout the experiments. 
We then conduct software-based ablation studies to quantify the impact of key design choices, including model capacity, graph construction, and filtration parameters, and to identify a configuration suitable for the subsequent hardware implementation.

\subsection{Dataset generation}
\label{subsec:implementation_details}

To train the proposed GNN, we utilized the Google Speech Commands v2 (GSCv2) dataset \cite{warden2018}, comprising over 100,000 one-second utterances. To capture the precise noise characteristics and temporal dynamics of the sensor, we generated a neuromorphic version of the dataset by processing the full audio corpus through the physical NAS hardware rather than relying on software simulations.

The acquisition pipeline involved streaming GSCv2 audio samples via an I2S codec to the NAS, which decomposed the signal into asynchronous AER events. These events were captured by an Opal Kelly XEM6310 FPGA module. Utilizing the FrontPanel SDK, we established a high-bandwidth USB 3.0 communication link between the FPGA and a host PC. The FPGA logic was programmed to append microsecond-resolution timestamps to the incoming AER packets, producing $(t, c, p)$ tuples. A dedicated Python script automated the process, synchronizing audio playback with event acquisition and converting the stream into .aedat format.

The generated datasets were approximately 15–-30\% longer than the original audio corpus. This extension resulted from intentional padding introduced during the hardware recording process, comprising pre-sample delays for error-free sequential acquisition, and post-sample delays to avoid positioning keywords at the end of the recording window, which degraded learning process.



\subsection{Implementation details}
\label{subsec:implementation_details}

\subsubsection{Training Setting}
\label{subsec:training_setting}

The keyword spotting system was trained for 50 epochs in floating point (FP32) using the Adam optimizer \cite{kingma2014adam} with a learning rate of $1e\text{-}3$ and weight decay of $1e\text{-}4$, together with a cosine-annealing learning-rate schedule \cite{loshchilov2016sgdr}. 
The network was then fine-tuned for 5 additional epochs using 8-bit quantization-aware training (QAT) with a constant learning rate of $1e\text{-}4$. 
All experiments used a batch size of 16 and were run on an NVIDIA GH200 GPU. 
The model was implemented using PyTorch \cite{Ansel_PyTorch_2_Faster_2024}. Checkpoints were selected based on the minimum validation loss and evaluated on the test set.







\subsection{Ablation studies}
\label{subsec:ablation_studies}

We perform ablation studies to quantify the impact of key design choices on keyword spotting performance and to select a configuration suitable for hardware implementation. 
We consider three factors: model size, graph-generation settings, and filtration parameters. 
Unless stated otherwise, results are obtained on the test set (64-cascade configuration) using the baseline configuration: 72 channels per layer; channel radius ($R_c$) 20 with skip step 2; low/high time radius ($R_t^{low/high}$) of 2000/10000; and filtration parameters with division factor 8, weight 32 and thresholds exponentially spaced from 64 to 32.

We evaluate both keyword classification and end-of-word temporal localization. We report classification accuracy (Acc.) and macro F1, as well as timestamp-conditioned accuracy (Ts-acc$_k$), which counts a prediction as correct only if the class matches and the predicted end-of-word time falls within $\pm k$ bins (10\,ms each). For the filtration ablation, we additionally report event-rate statistics.
All results are in FP32 (quantized models are evaluated in Section~\ref{subsec:HW}). Formal definition of all metrics, complete sweep results and model weights are released as open-source materials\footnote{\url{https://github.com/vision-agh/NAS-GNN-KWS}}.




\subsubsection{Model size}

We vary only the number of channels per layer in the set $\{$18, 36, 54, 72, 90, 108, 126$\}$.
Performance improves with increasing width, reaching the best results at 126 channels per layer (79.45\% accuracy, 65.51\% F1). 
However, gains beyond 72 channels are modest: the 72-channel model achieves 78.70\% accuracy and 62.29\% F1, while being substantially smaller (59.84k parameters vs. 179.56k for 126 channels). 
We therefore select 72 channels for hardware implementation.

\subsubsection{Graph generation}
We next examine the effect of graph-generation parameters.

\begin{figure}[!t]
    \centering
    \resizebox{\columnwidth}{!}{
    \includegraphics{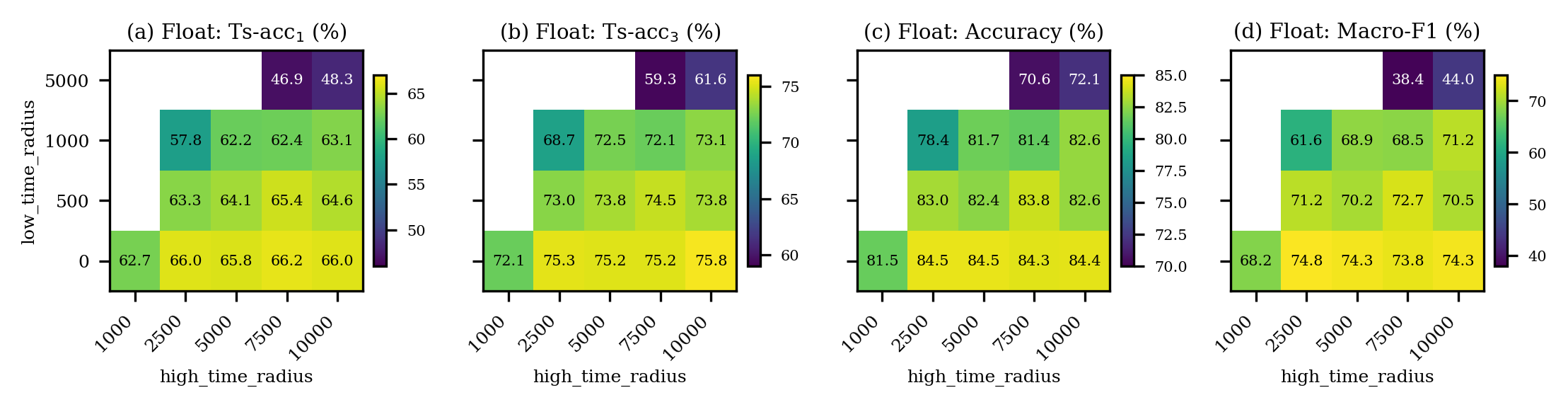}
    }
    \caption{Influence of the low and high time radius on the keyword-spotting metrics.}
    \label{fig:ablation_time_radius}
\end{figure}

\textit{- Time radius.}
Figure~\ref{fig:ablation_time_radius} shows the effect of varying the lower time radius $R_t^{low}$ (0--5000) and upper time radius $R_t^{high}$ (1000--10000).~Increasing $R_t^{low}$ consistently degrades all metrics: relative to $R_t^{low}=0$, setting it to 500 or 1000 decreases the average F1 by 1.93\% and 5.53\%, respectively.
The best results are achieved with $R_t^{low}=0$ and $R_t^{high}\in\{2500, 5000\}$ (84.5\% accuracy). We set $R_t^{low/high}=0/5000$ for subsequent experiments.


\textit{- Channel radius and skip step.}
Table~\ref{tab:ablation_graph_and_filtration}a reports performance for different channel-radius/skip-step pairs, with $R_t^{low/high}$ fixed to 0/5000.
We consider paired settings (10/1, 20/2, 30/3, 40/4, 50/5) that keep the maximum number of neighbours constant (20 plus a self-loop), alongside a baseline with $R_c=20$ and skip step 1 (dense neighbourhood). Overall, increasing either the radius or the skip step reduces performance.
The best configuration is $R_c=10$ with skip step 1 (85.89\% accuracy, 76.99\% F1). 
Together with the time-radius results, this suggests that the model benefits from a restricted spatiotemporal context, as broader neighbourhoods incorporate more irrelevant events.

\subsubsection{Filtration}
\label{subsec:filtration_ablation}

We evaluate the division factor and the threshold configuration, while keeping the weight parameter constant and setting the $R_t^{low/high}$ to 0/5000.

\textit{- Division factor.}
Table~\ref{tab:ablation_graph_and_filtration}b shows the effect of the division factor. 
Disabling filtration (--) substantially degrades performance (74.36\% accuracy) and produces the highest event rate (24.63 avg. kEv/s). 
Introducing filtration improves all metrics while reducing the event count. 
As the division factor increases from 6 to 10, accuracy improves (83.59\%$\to$84.93\%) while Ts-acc changes only marginally, indicating a trade-off between noise suppression and removal of informative events.
We select division factor 8, which achieves 84.46\% accuracy with an approximately twofold reduction in event rate.

\begin{table}[!t]
\centering
\caption{Ablation results for (a)~channel radius\,/\,skip step and (b)~filtration division factor. Bold marks the selected configuration; ``--''~denotes no filtration.}
\label{tab:ablation_graph_and_filtration}
 
\begin{minipage}[t]{0.47\columnwidth}
\centering
\scriptsize
\textbf{(a)} $R_t^{low/high}\!=\!0/5000$
 
\vspace{0.5ex}
\setlength{\tabcolsep}{7pt}
\begin{tabular}{@{}ccccc@{}}
\toprule
\makecell{$R_c$\,/\\[-1pt]skip} &
\makecell{Ts-\\[-1pt]acc$_1$} &
\makecell{Ts-\\[-1pt]acc$_3$} &
\makecell{Acc.\\[-1pt](\%)} &
\makecell{F1\\[-1pt](\%)} \\ \midrule
\textbf{10\,/\,1} & \textbf{67.4} & \textbf{77.0} & \textbf{85.9} & \textbf{77.0} \\
20\,/\,1 & 66.6 & 76.0 & 85.0 & 74.9 \\
20\,/\,2 & 58.3 & 68.6 & 78.7 & 62.3 \\
30\,/\,3 & 63.4 & 72.9 & 82.0 & 69.5 \\
40\,/\,4 & 63.5 & 72.7 & 82.1 & 70.2 \\
50\,/\,5 & 63.8 & 72.9 & 81.7 & 69.0 \\ \bottomrule
\end{tabular}
\end{minipage}%
\hfill
\begin{minipage}[t]{0.52\columnwidth}
\centering
\scriptsize
\textbf{(b)} $R_c\!=\!10$, skip\,$1$, $R_t^{low/high}\!=\!0/5000$
 
\vspace{0.5ex}
\setlength{\tabcolsep}{7pt}
\begin{tabular}{@{}cccccc@{}}
\toprule
\makecell{Div.\\[-1pt]fact.} &
\makecell{Ts-\\[-1pt]acc$_1$} &
\makecell{Ts-\\[-1pt]acc$_3$} &
\makecell{Acc.\\[-1pt](\%)} &
\makecell{F1\\[-1pt](\%)} &
\makecell{Avg.\\[-1pt]kEv/s} \\ \midrule
--  & 51.2 & 62.8 & 74.4 & 49.1 & 24.6 \\
6   & 66.5 & 76.0 & 83.6 & 72.6 & 7.9  \\
7   & 65.5 & 75.2 & 83.4 & 72.3 & 9.9  \\
\textbf{8} & \textbf{65.8} & \textbf{75.2} & \textbf{84.5} & \textbf{74.3} & \textbf{11.2} \\
9   & 66.2 & 75.9 & 84.7 & 74.6 & 11.8 \\
10  & 66.1 & 76.0 & 84.9 & 75.2 & 12.1 \\ \bottomrule
\end{tabular}
\end{minipage}
\end{table}

\textit{- Threshold configuration.}
We compare exponential, linear, and constant threshold schedules (weight 32, division factor 8).
Note that thresholds $\le w$ effectively disable filtration (e.g.\ the constant-32 setting). 
Among all tested schedules (detailed results in the additional material), linear thresholds of 48$\to$16 achieve the best performance (85.22\% accuracy, 75.61\% F1), slightly outperforming constant 48 (84.29\% accuracy).
This indicates that stronger filtration is beneficial for lower channels (higher frequency), whereas higher channels (lower frequency) require less suppression.

\subsubsection{Best configurations}

Based on the ablation studies, we select one configuration for each dataset variant and report the corresponding metrics in Table~\ref{tab:all_best_configs}. Across all selected configurations, the time radius, division factor, and threshold settings remain identical; only the channel radius and skip step differ. The best results are achieved by the parallel configurations, which generate more than twice as many events on average as the cascade configurations. Moreover, the frequency-dependent latency introduced by the cascade configuration (see Section~\ref{subsec:sensor}) hinders precise end-of-word detection, negatively affecting metrics that explicitly account for this parameter (Ts-acc$_1$, Ts-acc$_3$).

\begin{table}[!t]
\centering
\caption{Best ablation-selected configurations for cascade and parallel variants across datasets, reporting configuration, performance and throughput.}
\label{tab:all_best_configs}
\small
\setlength{\tabcolsep}{3pt}
\renewcommand{\arraystretch}{1.15}

\sisetup{
  detect-all,
  table-number-alignment=center
}

\resizebox{\columnwidth}{!}{
\begin{tabular}{l
S[table-format=2.0]
S[table-format=1.0]
S[table-format=4.0]
S[table-format=4.0]
S[table-format=1.0]
l
S[table-format=2.2]
S[table-format=2.2]
S[table-format=2.2]
S[table-format=2.2]
S[table-format=3.2]
S[table-format=2.2]
S[table-format=2.2]
}
\toprule
& \multicolumn{6}{c}{Configuration} & \multicolumn{4}{c}{Performance (\%)} & \multicolumn{3}{c}{Throughput} \\
\cmidrule(lr){2-7}\cmidrule(lr){8-11}\cmidrule(lr){12-14}

Dataset
& \multicolumn{1}{c}{$R_{c}$}
& \multicolumn{1}{c}{\shortstack{Skip\\step}}
& \multicolumn{1}{c}{\shortstack{$R_{\mathrm{t}}^{low}$}}
& \multicolumn{1}{c}{\shortstack{$R_{\mathrm{t}}^{high}$}}
& \multicolumn{1}{c}{\shortstack{div.\\factor}}
& \multicolumn{1}{c}{Thresholds}
& \multicolumn{1}{c}{\shortstack{Ts-acc$_1$}}
& \multicolumn{1}{c}{\shortstack{Ts-acc$_3$}}
& \multicolumn{1}{c}{Acc.}
& \multicolumn{1}{c}{F1}
& \multicolumn{1}{c}{\shortstack{Max.\\kEv./s}}
& \multicolumn{1}{c}{\shortstack{Avg.\\kEv./s}}
& \multicolumn{1}{c}{\shortstack{Avg.\\edges/Ev}}
\\
\midrule
32-cascade    & 5  & 1 & 0 & 5000 & 8 & exp.: {$64\rightarrow 32$} & 67.02 & 76.68 & 84.28 & 75.55 & 76.28  & 7.88  & 10.53 \\
32-parallel  & 5  & 1 & 0 & 5000 & 8 & exp.: {$64\rightarrow 32$} & 68.57 & 75.58 & 88.03 & 79.36 & 92.87  & 18.68 & 9.05  \\
64-cascade    & 10 & 1 & 0 & 5000 & 8 & exp.: {$64\rightarrow 32$} & 67.41 & 77.01 & 85.89 & 76.99 & 99.45  & 14.33 & 18.77 \\
64-parallel  & 10 & 1 & 0 & 5000 & 8 & exp.: {$64\rightarrow 32$} & 75.12 & 81.85 & 87.87 & 80.29 & 147.69 & 36.33 & 17.61 \\
128-cascade   & 20 & 2 & 0 & 5000 & 8 & exp.: {$64\rightarrow 32$} & 66.34 & 75.52 & 85.13 & 75.63 & 119.64 & 27.55 & 19.62 \\
128-parallel & 20 & 2 & 0 & 5000 & 8 & exp.: {$64\rightarrow 32$}
& 67.94
& 76.29
& 83.15
& 78.54
& 258.76
& 43.74
& 18.05
\\
\bottomrule
\end{tabular}
}
\end{table}


\subsubsection{Comparison with state-of-the-art}

To contextualize our results, we compare the best-performing configuration identified in the ablation studies with previously reported state-of-the-art results.
For consistency, we use \emph{accuracy} as the primary metric and restrict the comparison to works reporting accuracy on the SSC dataset, which is the closest match to our dataset.
In addition, we report the number of parameters as an indicator of model size (see Table~\ref{tab:sota_ssc}).

Following the categorization used above, \cite{dampfhoffer2022,bittar2022,Sadovsky2023,hammouamri2023,malettira2024,baronig2025,sun2025} report SNN-based models, \cite{schone2024,huber2024scaling} report state-space models; and \cite{Jeziorek2026HAGNN} report a graph-based approach.
For our method, we report results for the best-performing configuration within each NAS setting (32, 64, and 128 channels) and for both cascade and parallel variants, where available.

Overall, our approach achieves higher accuracy than almost all previously reported methods, including all SNN-based models and the prior graph-based implementation on SoC FPGA, while using substantially fewer parameters (59.84k). The only exception is Event-SSM~\cite{schone2024}, which reports a slightly higher peak accuracy (88.4\%) than our best configuration (32-parallel, 88.03\%). Importantly, our model is considerably smaller and is designed for end-to-end deployment on SoC FPGA, making it well suited for full hardware implementation under tight resource constraints.

\begin{table}[!t]
\centering
\caption{Accuracy and parameter-count comparison. Results for prior work are reported on the SSC dataset, while our models are evaluated on our dataset. Ranges indicate results reported for multiple model settings in the cited work.}
\label{tab:sota_ssc}
\small
\setlength{\tabcolsep}{5pt}
\resizebox{\columnwidth}{!}{
\begin{tabular}{@{}lllc c@{}}
\toprule
\textbf{Category} & \textbf{Reference} & \textbf{Model} & \textbf{\#Params} & \textbf{Acc. (\%)} \\
\midrule
\multirow{7}{*}{\emph{SNN}} & Dampfhoffer et al.~\cite{dampfhoffer2022} & SpikGRU                            & 280k      & 77.0 \\
& Bittar et al.~\cite{bittar2022}           & Recurrent SNN                      & 3.9M      & 77.4 \\
& Sadovsky et al.~\cite{Sadovsky2023}       & SNN-CNN                            & --        & 72.0 \\
& Hammouamri et al.~\cite{hammouamri2023}   & DCLS-Delays                        & 0.7--2.5M & 79.8--80.7 \\
& Malettira et al.~\cite{malettira2024}     & Temporal skips with delay learning & 1.4M      & 80.2 \\
& Baronig et al.~\cite{baronig2025}         & SE-adLIF                           & 1.6M      & 80.4 \\
& Sun et al.~\cite{sun2025}                 & PfA SNN                            & 0.1/0.7M  & 77.4/80.2 \\
\midrule
\multirow{2}{*}{\emph{SSM}} & Schone et al.~\cite{schone2024}           & Event-State-Space Model                          & 0.1/0.6M  & 85.3/88.4 \\
& Huber et al.~\cite{huber2024scaling}      & S5-RF                              & 1.8M      & 78.8 \\
\midrule
\multirow{1}{*}{\emph{GNN}} & Jeziorek et al.~\cite{Jeziorek2026HAGNN}  & Spectro-temporal graph on SoC FPGA & 8.6k--272k & 78.4--84.3 \\
\midrule
\multirow{6}{*}{\emph{\textbf{Our}}} & 32-cascade    & \multirow{6}{*}{End-to-end GNN with NAS on FPGA} & 59.84k & 84.61 \\
& 32-parallel  &  & 59.84k & 88.03 \\
& 64-cascade    &  & 59.84k & 85.89 \\
& 64-parallel  &  & 59.84k & 87.87 \\
& 128-cascade   &  & 59.84k & 85.03 \\
& 128-parallel &  & 59.84k & 83.15  \\
\bottomrule
\end{tabular}
}
\end{table}

\subsection{Hardware implementation}
\label{subsec:HW}

Based on the ablation studies (Section~\ref{subsec:ablation_studies}) and the system requirements (Section~\ref{subsec:gnn_model}), we developed a hardware implementation integrating both the NAS and the GNN on the same device, building upon the modules provided in~\cite{Jeziorek2026HAGNN}.

The graph neural network module supports all considered NAS configurations (32, 64, or 128 channels, in both parallel and serial variants). Due to efficient memory organization for storing features of previously processed events—required for the \texttt{PointNetConv} operation—BRAM usage remains independent of the channel count. Each memory entry stores nine 8-bit features (72 bits in total), resulting in a required depth of 1024 for the 128-channel configuration.  

Based on measured event statistics, we selected the number of parallel multipliers per graph convolution to process two 72-element feature vectors concurrently. With this configuration, a new event can be accepted every 0.47\,\textmu s to 4.07\,\textmu s (depending on the number of edges) at 200\,MHz. The resulting throughput is estimated between 2.1\,MEv./s and 245\,kEv./s, significantly exceeding the measured event rates for all considered datasets (see Table~\ref{tab:all_best_configs}).  
To handle short bursts of higher activity, we introduced a FIFO buffer between the filtering and graph generation modules.




\begin{table}[t]
\centering
\setlength{\tabcolsep}{10pt}
\caption{Evaluation of proposed design compared with GNN from \cite{Jeziorek2026HAGNN}.}
\label{tab:hw_comparison}
\begin{tabular}{l c c c}
\hline\hline
\multirow{2}{*}{\textbf{Metric}} & \textbf{GNN} & \textbf{64-parallel}  & \textbf{32-parallel}  \\
 & \textbf{\cite{Jeziorek2026HAGNN}} & \textbf{This work} & \textbf{This work} \\
\hline\hline
LUT & 125,130 & 88,370 & 82,739 \\
FF & 82,372 & 84,031 & 77,457\\
BRAM & 75.5 & 55.5 & 55.5\\
DSP & 140 & 83 & 83\\
Latency [\textmu s] & 10.53 & 25 (42) & 25 (35) \\
Throughput [keps] & 555 & 245 & 440 \\
Power  [W] & 1.18 & 1.16 & 1.12 \\
Accuracy & 73.5\% & 86.9\%  & 87.43\% \\
\hline
\end{tabular}
\end{table}

The complete system was implemented on AMD US+ ZCU104 and validated against the software reference model. For evaluation purposes, we selected the 32-channel and 64-channel parallel NAS configurations due to their high accuracy of 87.43\% and 86.9\% respectively (after 8-bit quantization). An additional advantage of the parallel variants is the elimination of additional low-frequency latency effects that are characteristic of the cascaded configuration. Both implementations were evaluated with respect to the following metrics (Table \ref{tab:hw_comparison}):

\begin{itemize}
    \item \textbf{Resource utilization} The proposed optimizations enabled a significant reduction in resource utilization compared to the GNN proposed in~\cite{Jeziorek2026HAGNN}. This resource efficiency allowed the full system to be deployed on the target platform without timing violations at a 200\,MHz clock frequency.
    \item \textbf{Latency}. Due to the MaxPool architecture, the system generates predictions every 10\,ms, invoking the network head once per window. With the timestamp propagation mechanism (see Section~\ref{subsec:gnn_model}), the effective GNN latency—defined as the time between the end of a 10\,ms window and the prediction—is typically equal to the head latency, i.e., 2.11\,\textmu s. In the worst case (when the last event occurs near the very end of the time window), it increases to a maximum of 18.62\,\textmu s.  The NAS exhibits an input-to-event latency of 23\,\textmu s resulting in overall end-to-end latency between 25 and 42\,\textmu s (for 32-channels between 25 and 35\,\textmu s).
    \item \textbf{Power consumption}. Average power of the KWS modules was estimated in Vivado using post-implementation simulation toggle-rates under 40\,kEv/s load. For 64-channel coniguration (simulated with an average of 18 edges/Ev.) yielding 0.539\,W dynamic and 0.595\,W static, and for 32-channels (avg. 9.8 edges/Ev.) - 0.494\,W dynamic and 0.595\,W static. With NAS power reported as 29.7\,mW in \cite{jimenez2017binaural}, the total average power is estimated at 1.16\,W and 1.12\,W, for 64-channel and 32-channel configurations respectively.
\end{itemize}

To the best of our knowledge, this is the first system enabling fully integrated, end-to-end keyword spotting on event-based audio with both the sensor and neural network implemented on a single FPGA device. There are, however, recent works on efficient KWS implementations for FPGA platforms evaluated on the original Google Speech Commands Dataset (not the spiking one) that report higher accuracies, e.g., a binary neural network accelerator on a Xilinx VC707 achieving 97.29\% in 536 clock cycles~\cite{zhang2025high}, and a configurable temporal-efficient neural network reaching 95.36\%~\cite{he2022configurable}. 
However, both approaches rely on precomputed features rather than processing raw audio directly, which introduces additional preprocessing overhead, power consumption, and latency. Furthermore, their predictions are generated once per 1\,s input sample.

In contrast, our approach eliminates the need for explicit aggregation or conventional preprocessing stages, thereby avoiding the associated latency and computational overhead. Furthermore, we generate predictions every 10\,ms, significantly increasing the temporal resolution of keyword detection. This combination of raw audio handling, frequent prediction updates, and adaptive scheduling distinguishes our architecture from existing FPGA-based KWS systems.

\section{Summary}
\label{sec:summary}

This paper presents an end-to-end, real-time keyword spotting system implemented entirely on a single FPGA through tight integration of a Neuromorphic Auditory Sensor (NAS) with a graph neural network (GNN). We publicly release both software and hardware implementations, together with recorded event-based versions of the GSCD for multiple NAS configurations.  

We propose a hardware-friendly event filtering method that reduces the number of processed events by approximately 47\% while improving classification accuracy. Furthermore, we introduce GNN architectural modifications derived from extensive ablation studies to reduce resource utilization, latency, and power consumption while maintaining high throughput. On hardware, the selected 64-channel parallel NAS configuration achieves 87.43\% accuracy after 8-bit quantization. The system produces predictions every 10\,ms, with an effective post-window inference latency of 35\,\textmu s and low average power consumption at 1.12W.

The proposed design enables accurate and high-speed inference directly from raw audio on a single device, making it well suited for edge computing applications in mobile robotics and IoT. In future work, we plan to validate the proposed approach in real-world robotic human–machine interaction scenarios. Additionally, we intend to investigate the feasibility of an ASIC implementation to achieve an even more efficient and optimized solution.

\section{Acknowledgments} 

This work was supported by: the Polish National Science Centre projects 
no.~2024/\hspace{0pt}53/\hspace{0pt}N/\hspace{0pt}ST6/\hspace{0pt}04254 
and no.~2024/\hspace{0pt}53/\hspace{0pt}N/\hspace{0pt}ST6/\hspace{0pt}04331, the “Excellence initiative – research university” programme, 16.16.120.773 for the AGH University of Krakow. This work is a part of the projects PID2023-149071NB-C54 (NEKOR) funded by "Ministerio de Ciencia, Innovación y Universidades"/AEI/\hspace{0pt}10.13039/\hspace{0pt}501100011033, by “ERDF A way of making Europe” and by the European Union NextGenerationEU/PRTR, by grant~USECHIP (TSI-069100-2023-001), project funded by the Secretary of State for Telecommunications and Digital Infrastructure, Ministry for Digital Transformation and Civil Service and by the European Union–NextGenerationEU/PRTR.
We gratefully acknowledge Polish high-\hspace{0pt}performance computing infrastructure 
PLGrid (HPC~Center: ACK Cyfronet AGH) for providing computer facilities 
and support within computational grant 
no.~PLG/\hspace{0pt}2026/\hspace{0pt}019156.

%
%
%
\bibliographystyle{splncs04}
\bibliography{bibtex}

\begin{thebibliography}{10}
\providecommand{\url}[1]{\texttt{#1}}
\providecommand{\urlprefix}{URL }
\providecommand{\doi}[1]{https://doi.org/#1}

\bibitem{abbott1999lapicque}
Abbott, L.F.: Lapicque’s introduction of the integrate-and-fire model neuron
  (1907). Brain research bulletin  \textbf{50}(5-6),  303--304 (1999)

\bibitem{Ansel_PyTorch_2_Faster_2024}
Ansel, J., Yang, E., He, H., Gimelshein, N., Jain, A., Voznesensky, M., Bao,
  B., Bell, P., Berard, D., Burovski, E., Chauhan, G., Chourdia, A., Constable,
  W., Desmaison, A., DeVito, Z., Ellison, E., Feng, W., Gong, J., Gschwind, M.,
  Hirsh, B., Huang, S., Kalambarkar, K., Kirsch, L., Lazos, M., Lezcano, M.,
  Liang, Y., Liang, J., Lu, Y., Luk, C., Maher, B., Pan, Y., Puhrsch, C., Reso,
  M., Saroufim, M., Siraichi, M.Y., Suk, H., Suo, M., Tillet, P., Wang, E.,
  Wang, X., Wen, W., Zhang, S., Zhao, X., Zhou, K., Zou, R., Mathews, A.,
  Chanan, G., Wu, P., Chintala, S.: {PyTorch 2: Faster Machine Learning Through
  Dynamic Python Bytecode Transformation and Graph Compilation}. In: 29th ACM
  International Conference on Architectural Support for Programming Languages
  and Operating Systems, Volume 2 (ASPLOS '24). ACM (Apr 2024).
  \doi{10.1145/3620665.3640366},
  \url{https://docs.pytorch.org/assets/pytorch2-2.pdf}

\bibitem{baronig2025}
Baronig, M., Ferrand, R., Sabathiel, S., Legenstein, R.: Advancing
  spatio-temporal processing through adaptation in spiking neural networks.
  Nature Communications  \textbf{16}(1), ~5776 (Jul 2025).
  \doi{10.1038/s41467-025-60878-z},
  \url{https://doi.org/10.1038/s41467-025-60878-z}

\bibitem{bittar2022}
Bittar, A., Garner, P.N.: A surrogate gradient spiking baseline for speech
  command recognition. Frontiers in Neuroscience  \textbf{16} (2022).
  \doi{10.3389/fnins.2022.865897}

\bibitem{carpegna2024spiker}
Carpegna, A., Savino, A., Carlo, S.D.: {Spiker+: a framework for the generation
  of efficient Spiking Neural Networks FPGA accelerators for inference at the
  edge}. IEEE Transactions on Emerging Topics in Computing (01),  1--15 (Dec
  2024). \doi{10.1109/TETC.2024.3511676}

\bibitem{chan2007aer}
Chan, V., Liu, S.C., van Schaik, A.: Aer ear: A matched silicon cochlea pair
  with address event representation interface. IEEE Transactions on Circuits
  and Systems I: Regular Papers  \textbf{54}(1),  48--59 (2007).
  \doi{10.1109/TCSI.2006.887979}

\bibitem{cramer_heidelberg_2020}
Cramer, B., Stradmann, Y., Schemmel, J., Zenke, F.: The {Heidelberg} {Spiking}
  {Data} {Sets} for the {Systematic} {Evaluation} of {Spiking} {Neural}
  {Networks}. IEEE Transactions on Neural Networks and Learning Systems pp.
  1--14 (2020). \doi{10.1109/TNNLS.2020.3044364}

\bibitem{cramer2020}
Cramer, B., Stradmann, Y., Schemmel, J., Zenke, F.: The heidelberg spiking data
  sets for the systematic evaluation of spiking neural networks. IEEE
  Transactions on Neural Networks and Learning Systems  \textbf{33}(7),
  2744--2757 (2022). \doi{10.1109/TNNLS.2020.3044364},
  \url{https://zenkelab.org/resources/spiking-heidelberg-datasets-shd/}

\bibitem{dampfhoffer2022}
Dampfhoffer, M., Mesquida, T., Valentian, A., Anghel, L.: Investigating
  current-based and gating approaches for accurate and energy-efficient
  spiking recurrent neural networks. In: Pimenidis, E., Angelov, P., Jayne, C.,
  Papaleonidas, A., Aydin, M. (eds.) Artificial Neural Networks and Machine
  Learning -- ICANN 2022. pp. 359--370. Springer Nature Switzerland, Cham
  (2022). \doi{10.1007/978-3-031-15934-3_30}

\bibitem{gambin2010digital}
Gambin, I., Grech, I., Casha, O., Gatt, E., Micallef, J.: Digital cochlea model
  implementation using xilinx xc3s500e spartan-3e fpga. In: 2010 17th IEEE
  International Conference on Electronics, Circuits and Systems. pp. 946--949
  (2010). \doi{10.1109/ICECS.2010.5724669}

\bibitem{hammouamri2023}
Hammouamri, I., Khalfaoui-Hassani, I., Masquelier, T.: Learning delays in
  spiking neural networks using dilated convolutions with learnable spacings
  (2023), \url{https://arxiv.org/abs/2306.17670}

\bibitem{he2022configurable}
He, K., Chen, D., Su, T.: A configurable accelerator for keyword spotting based
  on small-footprint temporal efficient neural network. Electronics
  \textbf{11}(16), ~2571 (2022)

\bibitem{huber2024scaling}
Huber, T.E., Lecomte, J., Polovnikov, B., von Arnim, A.: Scaling up
  resonate-and-fire networks for fast deep learning. In: Del~Bue, A., Canton,
  C., Pont-Tuset, J., Tommasi, T. (eds.) Computer Vision -- ECCV 2024
  Workshops. pp. 241--258. Springer Nature Switzerland, Cham (2025).
  \doi{10.1007/978-3-031-92460-6_15}

\bibitem{Jeziorek2026HAGNN}
Jeziorek, K., Wzorek, P., Blachut, K., Nakano, H., Dampfhoffer, M., Mesquida,
  T., Nishi, H., Dalgaty, T., Kryjak, T.: Hardware-accelerated graph neural
  networks: an alternative approach for neuromorphic event-based audio
  classification and keyword spotting on soc fpga (2026),
  \url{https://arxiv.org/abs/2602.16442}

\bibitem{5596845}
Jimenez-Fernandez, A., Linares-Barranco, A., Paz-Vicente, R., Jiménez, G.,
  Civit, A.: Building blocks for spikes signals processing. In: The 2010
  International Joint Conference on Neural Networks (IJCNN). pp.~1--8 (2010).
  \doi{10.1109/IJCNN.2010.5596845}

\bibitem{jimenez2017binaural}
Jim{\'e}nez-Fern{\'a}ndez, A., Cerezuela-Escudero, E., Mir{\'o}-Amarante, L.,
  Dom{\'\i}nguez-Morales, M.J., Gomez-Rodriguez, F., Linares-Barranco, A.,
  Jim{\'e}nez-Moreno, G.: A binaural neuromorphic auditory sensor for fpga: A
  spike signal processing approach. IEEE Trans. Neural Netw. Learning Syst.
  \textbf{28}(4),  804--818 (2017)

\bibitem{kingma2014adam}
Kingma, D.P., Ba, J.: Adam: A method for stochastic optimization. arXiv
  preprint arXiv:1412.6980  (2014)

\bibitem{DAS_arch}
Liu, S.C., van Schaik, A., Minch, B.A., Delbruck, T.: Asynchronous binaural
  spatial audition sensor with 2$\,\times\,$64$\,\times\,$4 channel output.
  IEEE Transactions on Biomedical Circuits and Systems  \textbf{8}(4),
  453--464 (2014). \doi{10.1109/TBCAS.2013.2281834}

\bibitem{loshchilov2016sgdr}
Loshchilov, I., Hutter, F.: Sgdr: Stochastic gradient descent with warm
  restarts. arXiv preprint arXiv:1608.03983  (2016)

\bibitem{lyonandmead}
Lyon, R., Mead, C.: An analog electronic cochlea. IEEE Transactions on
  Acoustics, Speech, and Signal Processing  \textbf{36}(7),  1119--1134 (1988).
  \doi{10.1109/29.1639}

\bibitem{malettira2024}
Malettira, P.G., Negi, S., Ponghiran, W., Roy, K.: {TSkips: Efficiency Through
  Explicit Temporal Delay Connections in Spiking Neural Networks} (2024),
  \url{https://arxiv.org/abs/2411.16711}

\bibitem{quantisenc2024}
Matinizadeh, S., Pacik-Nelson, N., Polykretis, I., Tishbi, K., Kumar, S.,
  Varshika, M.L., Mohammadhassani, A., Mishra, A., Kandasamy, N., Shackleford,
  J., Gallo, E., Das, A.: A fully-configurable open-source software-defined
  digital quantized spiking neural core architecture (2024),
  \url{https://arxiv.org/abs/2404.02248}

\bibitem{nakano2025hardware}
Nakano, H., Blachut, K., Jeziorek, K., Wzorek, P., Dampfhoffer, M., Mesquida,
  T., Nishi, H., Kryjak, T., Dalgaty, T.: Hardware-accelerated event-graph
  neural networks for low-latency time-series classification on soc fpga. In:
  International Symposium on Applied Reconfigurable Computing. pp. 51--68.
  Springer (2025)

\bibitem{perez2021}
Perez-Nieves, N., Leung, V.C.H., Dragotti, P.L., Goodman, D.F.M.: Neural
  heterogeneity promotes robust learning. Nature Communications
  \textbf{12}(1), ~5791 (Oct 2021). \doi{10.1038/s41467-021-26022-3},
  \url{https://doi.org/10.1038/s41467-021-26022-3}

\bibitem{qi2017pointnet}
Qi, C.R., Yi, L., Su, H., Guibas, L.J.: Pointnet++: Deep hierarchical feature
  learning on point sets in a metric space (2017),
  \url{https://arxiv.org/abs/1706.02413}

\bibitem{lars}
Rafeldt, L., Mesquida, T., Nakano, H., Dampfhoffer, M., Moro, F., Vivet, P.,
  Payvand, M., Dalgaty, T.: Event-based audio prediction with spectro-temporal
  event-graphs. In: 2025 IEEE International Symposium on Circuits and Systems
  (ISCAS). pp.~1--5 (2025). \doi{10.1109/ISCAS56072.2025.11043865}

\bibitem{rossbroich2022}
Rossbroich, J., Gygax, J., Zenke, F.: Fluctuation-driven initialization for
  spiking neural network training. Neuromorphic Computing and Engineering
  \textbf{2}(4),  044016 (Dec 2022). \doi{10.1088/2634-4386/ac97bb}

\bibitem{Sadovsky2023}
Sadovsky, E., Jakubec, M., Jarina, R.: Speech command recognition based on
  convolutional spiking neural networks. In: 2023 33rd International Conference
  Radioelektronika (RADIOELEKTRONIKA). pp.~1--5 (2023).
  \doi{10.1109/RADIOELEKTRONIKA57919.2023.10109082}

\bibitem{schone2024}
Schöne, M., Sushma, N.M., Zhuge, J., Mayr, C., Subramoney, A., Kappel, D.:
  Scalable event-by-event processing of neuromorphic sensory signals with deep
  state-space models. In: 2024 International Conference on Neuromorphic Systems
  (ICONS). pp. 124--131 (2024). \doi{10.1109/ICONS62911.2024.00026}

\bibitem{sun2025}
Sun, P., Wu, J., Devos, P., Botteldooren, D.: Towards parameter-free
  attentional spiking neural networks. Neural Networks  \textbf{185},  107154
  (2025). \doi{https://doi.org/10.1016/j.neunet.2025.107154},
  \url{https://www.sciencedirect.com/science/article/pii/S0893608025000334}

\bibitem{thakur2014fpga}
Thakur, C.S., Hamilton, T.J., Tapson, J., van Schaik, A., Lyon, R.F.: Fpga
  implementation of the car model of the cochlea. In: 2014 IEEE International
  Symposium on Circuits and Systems (ISCAS). pp. 1853--1856 (2014).
  \doi{10.1109/ISCAS.2014.6865519}

\bibitem{wang2015design}
Wang, S., Koickal, T.J., Enemali, G., Gouveia, L., Wang, L., Hamilton, A.:
  Design of a silicon cochlea system with biologically faithful response. In:
  2015 International Joint Conference on Neural Networks (IJCNN). pp.~1--7
  (2015). \doi{10.1109/IJCNN.2015.7280828}

\bibitem{warden2018speech}
Warden, P.: Speech commands: A dataset for limited-vocabulary speech
  recognition. arXiv preprint arXiv:1804.03209  (2018)

\bibitem{warden2018}
Warden, P.: Speech commands: A dataset for limited-vocabulary speech
  recognition (2018), \url{https://arxiv.org/abs/1804.03209}

\bibitem{Yin2021}
Yin, B., Corradi, F., Boht{\'e}, S.M.: Accurate and efficient time-domain
  classification with adaptive spiking recurrent neural networks. Nature
  Machine Intelligence  \textbf{3}(10),  905--913 (Oct 2021).
  \doi{10.1038/s42256-021-00397-w},
  \url{https://doi.org/10.1038/s42256-021-00397-w}

\bibitem{zhang2025high}
Zhang, A., Shi, J., Qian, H., Wang, J.: High precision speech keyword spotting
  based on binary deep neural network in fpga. Entropy  \textbf{27}(11), ~1143
  (2025)

\end{thebibliography}

\end{document}


\title{End-to-End Keyword Spotting on FPGA Using Graph Neural Networks with a Neuromorphic Auditory Sensor \newline (Supplementary Material)}

\author{Anonimized Authors}
\authorrunning{Anonimized Authors}

%
\maketitle              
\section{Results of filtration}

This section illustrates the effect of the proposed filtration on representative files from the generated dataset. Figure~\ref{fig:example_data_filtration1} shows an example from the 64-cascade dataset, whereas Figure~\ref{fig:example_data_filtration2} presents the corresponding example from the 64-parallel dataset. For the 64-cascade example, the total number of events decreases from 87.38~kEv. (unfiltered) to 42.39~kEv. after filtration.

\begin{figure}[!ht]
    \centering
    \begin{subfigure}[t]{\columnwidth}
        \centering
        \includegraphics[width=\columnwidth]{figures/data/event_graph_64_serial_unfiltered.png}
    \end{subfigure}
    \begin{subfigure}[t]{\columnwidth}
        \centering
        \includegraphics[width=\columnwidth]{figures/data/event_graph_64_serial_filtered.png}
    \end{subfigure}
    \caption{Example of the generated data from a sample without filtration (top) and with filtration (bottom) for 64-cascade configuration.}
    \label{fig:example_data_filtration1}
\end{figure}

\begin{figure}[!ht]
    \centering
    \begin{subfigure}[t]{\columnwidth}
        \centering
        \includegraphics[width=\columnwidth]{figures/data/event_graph_64_parallel_unfiltered.png}
    \end{subfigure}
    \begin{subfigure}[t]{\columnwidth}
        \centering
        \includegraphics[width=\columnwidth]{figures/data/event_graph_64_parallel_filtered.png}
    \end{subfigure}
    \caption{Example of the generated data from a sample without filtration (top) and with filtration (bottom) for 64-parallel configuration.}
    \label{fig:example_data_filtration2}
\end{figure}

\section{Metrics}
\label{sec:metrics}

We report complementary metrics that capture both classification performance and end-of-word temporal localization. 
Let $N$ denote the number of test samples. 
For each sample $i$, let $t_i$ be the ground-truth end-of-word timestamp (in time bins), and let $\hat{t}_i$ be the predicted end-of-word timestamp defined as the time bin with the maximum end-of-word confidence over the sample. 
Let $y_i$ denote the ground-truth class label, and let $\hat{y}_i$ denote the predicted class label produced by the classifier head at time bin $\hat{t}_i$.

\paragraph{- Accuracy (Acc. \%)} is the proportion of samples whose predicted class label matches the ground truth:
\begin{equation}
\mathrm{Acc} = \frac{100}{N} \sum_{i=1}^{N} \mathbf{1}\{\hat{y}_i = y_i\}.
\end{equation}

\paragraph{- Timestamp-conditioned accuracy (Ts-acc$_k$ \%)} is the fraction of samples that are both correctly classified and localized within a tolerance of $\pm k$ time bins, where each bin corresponds to $10~\mathrm{ms}$:
\begin{equation}
\mathrm{Ts\text{-}acc}_k =
\frac{100}{N} \sum_{i=1}^{N}
\mathbf{1}\{\hat{y}_i = y_i\}\,
\mathbf{1}\{|\hat{t}_i - t_i| \le k\}.
\end{equation}

\paragraph{- Macro F1 score (F1 \%)} is computed by first evaluating the per-class F1 score
\begin{equation}
\mathrm{F1}_c = \frac{2\,\mathrm{Prec}_c\,\mathrm{Rec}_c}{\mathrm{Prec}_c+\mathrm{Rec}_c+\varepsilon},
\end{equation}
and then averaging over the classes present in the ground truth:
\begin{equation}
\mathrm{F1}_{\mathrm{macro}} =
\frac{100}{|\mathcal{C}_p|} \sum_{c \in \mathcal{C}_p} \mathrm{F1}_c,
\qquad
\mathcal{C}_p=\{c \mid \mathrm{supp}(c)>0\},
\end{equation}
where $\mathrm{supp}(c)$ denotes the number of ground-truth samples in class $c$ and $\varepsilon = 10^{-12}$ is used for numerical stability.

\paragraph{- Event-rate statistics.}
To characterize the system output and quantify the effect of the filtration stage, we additionally report the mean and maximum number of NAS events observed within $10~\mathrm{ms}$ windows over the test set.

\section{Ablation studies}
\label{sec:ablation_studies}

In this section, we present extended results of the performed ablation studies, reporting each ablation in a table or heatmap and linking it to the corresponding model (will be added after acceptance).
We consider three factors: model size, graph-generation settings, and filtration parameters. 
Unless stated otherwise, results are obtained in FP32 on the test set using the default configuration: 72 channels per layer; channel radius 20 with skip step 2; low/high time radius of 2000/10000; and filtration parameters with division factor 8 and weight 32, with thresholds exponentially spaced from 64 to 32.

\subsubsection{Model size}

Table~\ref{tab:ablation_model_size} reports keyword-spotting performance (Ts-acc$_1$, Ts-acc$_3$, accuracy, and F1) and parameter count as the number of channels per layer is varied.

\begin{table}[!ht]
\centering
\caption{Ablation study of model size: keyword-spotting performance as a function of channels per layer with parameter count in (thousands).}
\label{tab:ablation_model_size}
\setlength{\tabcolsep}{2pt}
\resizebox{\textwidth}{!}{
\begin{tabular}{@{}ccccccc@{}}
\toprule
\textbf{Num. channels} & 
\textbf{Num. parameters (k)} & 
\multicolumn{1}{c}{\textbf{Ts-acc$_1$ (\%)}} & 
\multicolumn{1}{c}{\textbf{Ts-acc$_3$ (\%)}} & 
\multicolumn{1}{c}{\textbf{Acc. (\%)}} & 
\multicolumn{1}{c}{\textbf{F1 (\%)}} &
\textbf{Model} \\ \midrule
18  & 4.27   & 47.17\% & 57.81\% & 68.51\% & 29.05\% & \href{https://example.com/after/accept}{link} \\
36  & 15.67  & 56.44\% & 66.82\% & 76.30\% & 56.39\% & \href{https://example.com/after/accept}{link} \\
54  & 34.19  & 56.83\% & 67.62\% & 78.08\% & 60.74\% & \href{https://example.com/after/accept}{link} \\
72  & 59.84  & 58.28\% & 68.59\% & 78.70\% & 62.29\% & \href{https://example.com/after/accept}{link} \\
90  & 92.62  & 58.76\% & 68.83\% & 78.27\% & 63.86\% & \href{https://example.com/after/accept}{link} \\
108 & 132.53 & 57.72\% & 68.80\% & 78.74\% & 64.70\% & \href{https://example.com/after/accept}{link} \\
126 & 179.56 & 59.72\% & 70.02\% & 79.45\% & 65.51\% & \href{https://example.com/after/accept}{link} \\ \bottomrule
\end{tabular}
}
\end{table}

\subsubsection{Graph generation}

\begin{figure}[!t]
    \centering
    \resizebox{\columnwidth}{!}{
    \includegraphics{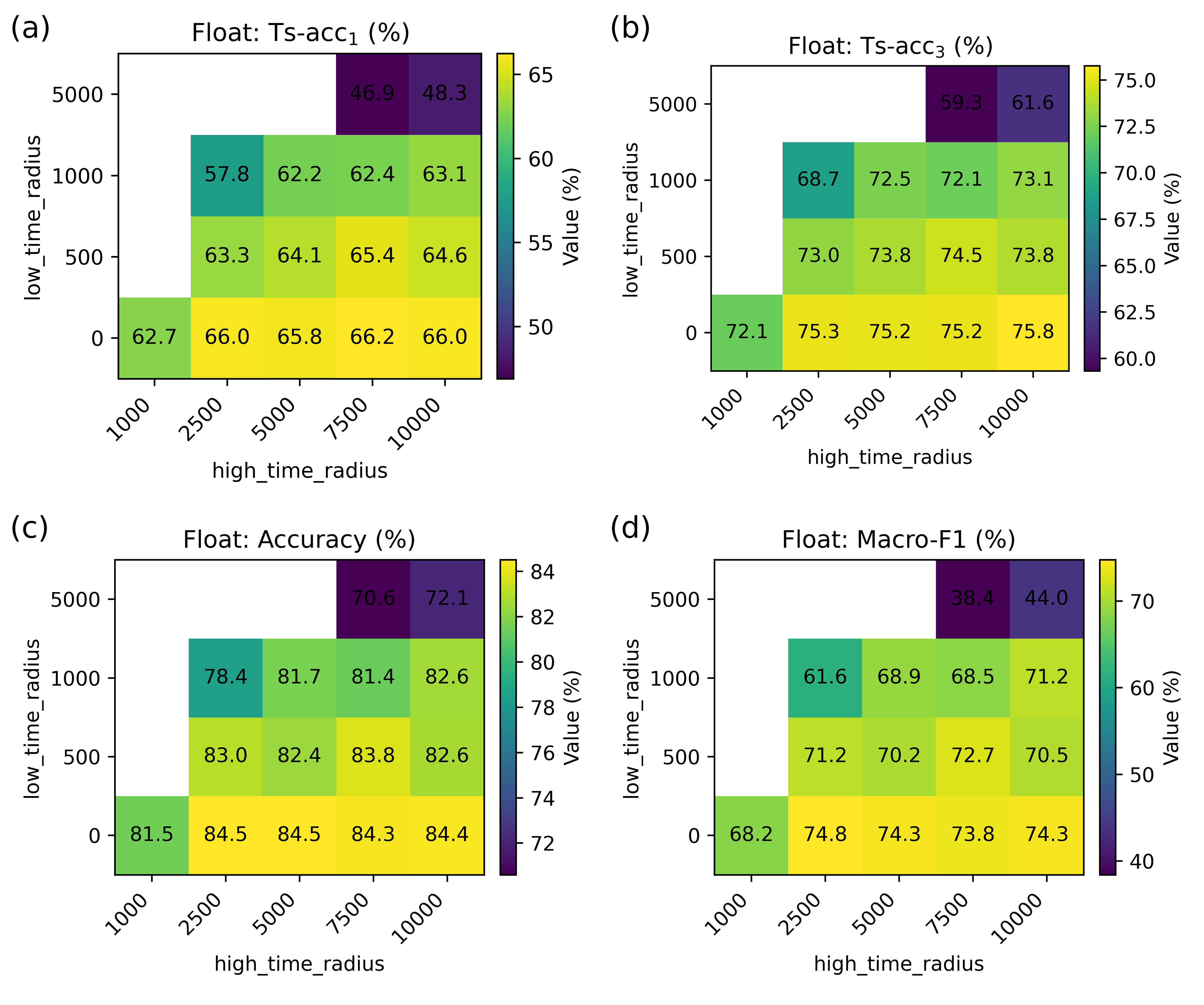}
    }
    \caption{Influence of the low and high time radius on the keyword-spotting metrics.}
    \label{fig:ablation_time_radius}
\end{figure}

\paragraph{Time radius.}
Figure~\ref{fig:ablation_time_radius} presents a heatmap of keyword-spotting performance metrics as a function of the low and high time radius.

\paragraph{Channel radius and skip step.}
Table~\ref{tab:ablation_channel_radius} reports keyword-spotting performance metrics for different channel-radius and skip-step configurations (with fixed low/high time radii), evaluating how changes in spatial connectivity affect performance.

\begin{table}[!ht]
\centering
\caption{Keyword-spotting metrics for different channel-radius/skip-step settings, with low time radius = 0 and high time radius = 5000.}
\label{tab:ablation_channel_radius}
\setlength{\tabcolsep}{6pt}
\resizebox{\columnwidth}{!}{
\begin{tabular}{@{}ccccccc@{}}
\toprule
\textbf{Ch. radius} & 
\textbf{Skip Step} & 
\multicolumn{1}{c}{\textbf{Ts-acc$_1$ (\%)}} & 
\multicolumn{1}{c}{\textbf{Ts-acc$_3$ (\%)}} & 
\multicolumn{1}{c}{\textbf{Acc. (\%)}} & 
\multicolumn{1}{c}{\textbf{F1 (\%)}} &
\textbf{Model} \\ \midrule
10 & 1 & 67.41\% & 77.01\% & 85.89\% & 76.99\% & \href{https://example.com/after/acceptance}{link} \\
20 & 1 & 66.55\% & 75.97\% & 84.95\% & 74.94\% & \href{https://example.com/after/acceptance}{link} \\
20 & 2 & 58.28\% & 68.59\% & 78.70\% & 62.29\% & \href{https://example.com/after/acceptance}{link} \\
30 & 3 & 63.40\% & 72.89\% & 81.96\% & 69.53\% & \href{https://example.com/after/acceptance}{link} \\
40 & 4 & 63.45\% & 72.70\% & 82.13\% & 70.16\% & \href{https://example.com/after/acceptance}{link} \\
50 & 5 & 63.76\% & 72.92\% & 81.68\% & 68.95\% & \href{https://example.com/after/acceptance}{link} \\ \bottomrule
\end{tabular}
}
\end{table}

\subsubsection{Filtration}
\label{subsec:filtration_ablation}

We evaluate the division factor and the threshold configuration, while keeping the weight parameter constant and setting the low/high time radius to 0/5000.

\paragraph{Division factor.}
Table~\ref{tab:ablation_filtration_decay} summarizes keyword-spotting performance metrics and event statistics (average and maximum events per 10,ms) as a function of the division factor, including the no-filtration baseline.

\begin{table}[!t]
\centering
\caption{Influence of the division factor on the metric of the keyword spotting. Value of the division factor equal to - means no filtration.}
\label{tab:ablation_filtration_decay}
\setlength{\tabcolsep}{2pt}
\resizebox{\columnwidth}{!}{
\begin{tabular}{@{}cccccccc@{}}
\toprule
\textbf{Div. factor} & 
\multicolumn{1}{c}{\textbf{Ts-acc$_1$ (\%)}} & 
\multicolumn{1}{c}{\textbf{Ts-acc$_3$ (\%)}} & 
\multicolumn{1}{c}{\textbf{Acc. (\%)}} & 
\multicolumn{1}{c}{\textbf{F1 (\%)}} & 
\multicolumn{1}{c}{\textbf{Avg. Ev./10ms}} & 
\multicolumn{1}{c}{\textbf{Max. Ev./10ms}} &
\textbf{Model} \\ \midrule
-  & 51.17\% & 62.79\% & 74.36\% & 49.14\% & 245.86 & 3293 & \href{https://example.com/after/acceptance}{link} \\
6  & 66.45\% & 76.02\% & 83.59\% & 72.56\% & 78.96 & 1599 & \href{https://example.com/after/acceptance}{link} \\
7  & 65.54\% & 75.16\% & 83.43\% & 72.29\% & 99.07 & 1629 & \href{https://example.com/after/acceptance}{link} \\
8  & 65.81\% & 75.16\% & 84.46\% & 74.29\% & 112.11 & 1633 & \href{https://example.com/after/acceptance}{link} \\
9  & 66.23\% & 75.86\% & 84.70\% & 74.64\% & 118.27 & 1640 & \href{https://example.com/after/acceptance}{link} \\
10 & 66.12\% & 75.96\% & 84.93\% & 75.17\% & 121.42 & 1644 & \href{https://example.com/after/acceptance}{link} \\ \bottomrule
\end{tabular}
}
\end{table}

\paragraph{Threshold configuration.}
Table~\ref{tab:ablation_filtration_thresholds_float} compares keyword spotting performance metrics and event statistics for different filtration-threshold generation methods (exponential, linear, and constant) and threshold ranges, including a no-filtration case when the threshold does not exceed the weight.

\begin{table}[!t]
\centering
\caption{Effect of filtration-threshold configuration on keyword-spotting performance. Bold values indicate the best configuration.}
\label{tab:ablation_filtration_thresholds_float}
\setlength{\tabcolsep}{4pt}
\resizebox{\columnwidth}{!}{
\begin{tabular}{@{}lcccccccc@{}}
\toprule
\textbf{Method} &
\textbf{Thresholds} &
\multicolumn{1}{c}{\textbf{Ts-acc$_1$ (\%)}} &
\multicolumn{1}{c}{\textbf{Ts-acc$_3$ (\%)}} &
\multicolumn{1}{c}{\textbf{Acc. (\%)}} &
\multicolumn{1}{c}{\textbf{F1 (\%)}} &
\multicolumn{1}{c}{\textbf{Avg. Ev./10ms}} &
\multicolumn{1}{c}{\textbf{Max. Ev./10ms}} &
\textbf{Model} \\ \midrule
exponential & 96$\to$64 & 62.44\% & 71.88\% & 82.37\% & 70.12\% & 74.36 & 1098 & \href{https://example.com/after/acceptance}{link} \\
exponential & 96$\to$48 & 65.24\% & 74.55\% & 83.77\% & 73.02\% & 76.40 & 1261 & \href{https://example.com/after/acceptance}{link} \\ 
exponential & 80$\to$48 & 62.65\% & 72.49\% & 82.40\% & 69.96\% & 82.27 & 1360 & \href{https://example.com/after/acceptance}{link} \\
exponential & 48$\to$16 & 64.92\% & 74.48\% & 83.71\% & 72.72\% & 169.70 & 2604 & \href{https://example.com/after/acceptance}{link} \\ \midrule
linear      & 96$\to$64 & 63.93\% & 74.31\% & 83.14\% & 72.06\% & 74.16 & 1101 & \href{https://example.com/after/acceptance}{link} \\
linear      & 96$\to$48 & 62.50\% & 72.47\% & 82.44\% & 70.23\% & 74.99 & 1209 & \href{https://example.com/after/acceptance}{link} \\ 
linear      & 80$\to$48 & 65.40\% & 74.55\% & 84.15\% & 73.50\% & 80.18 & 1330 & \href{https://example.com/after/acceptance}{link} \\
linear      & 64$\to$32 & 65.07\% & 74.92\% & 83.81\% & 72.96\% & 110.84 & 1630 & \href{https://example.com/after/acceptance}{link} \\
linear & 48$\to$16 & 66.45\% & 76.17\% & 85.22\% & 75.61\% & 145.70 & 2373 & \href{https://example.com/after/acceptance}{link} \\ \midrule
constant    & 96 & 61.85\% & 73.13\% & 81.72\% & 68.82\% & 64.63 & 1013 & \href{https://example.com/after/acceptance}{link} \\
constant    & 80 & 64.81\% & 75.56\% & 84.33\% & 73.95\% & 78.33 & 1101 & \href{https://example.com/after/acceptance}{link} \\
constant    & 64 & 63.79\% & 74.51\% & 83.53\% & 72.41\% & 83.61 & 1524 & \href{https://example.com/after/acceptance}{link} \\
constant    & 48 & 66.45\% & 75.49\% & 84.29\% & 73.85\% & 115.21 & 1627 & \href{https://example.com/after/acceptance}{link} \\
constant    & 32 & 51.17\% & 62.79\% & 74.36\% & 49.14\% & 245.864 & 3293 & \href{https://example.com/after/acceptance}{link} \\ \bottomrule
\end{tabular}
}
\end{table}

%
%
%